\documentclass[runningheads]{llncs}
\usepackage{graphicx}
\usepackage{listings}
\usepackage{algorithm2e}
\begin{document}
\title{Evolution of Ant Colony Optimization Algorithm --- A Brief Literature Review}
%
%
\author{Aleem Akhtar}

	\institute{SEECS-NUST, ISLAMABAD \\
		\email{aleem.akhtar@seecs.edu.pk} \\
		 }
%
%
%
\maketitle              
\begin{abstract}
Ant Colony Optimzation (ACO) is a metaheuristic proposed by Marco Dorigo in 1991 based on behavior of biological ants. Pheromone laying and selection of shortest route with the help of pheromone inspired development of first ACO algorithm. Since, presentation of first such algorithm, many researchers have worked and published their research in this field. Though initial results were not so promising but recent developments have made this metaheuristic a significant algorithm in Swarm Intelligence. This research presents a brief overview of recent developments carried out in ACO algorithms in terms of both applications and algorithmic developments. For application developments, multi-objective optimization, continuous optimization and time-varying NP-hard problems have been presented. While to review articles based on algorithmic development, hybridization and parallel architectures have been investigated.

\end{abstract}
\section{Introduction}
Ant Colony Optimization or simply ACO is a meateuristic that is used to solve optimization problems that are complex combinatorial ~\cite{dorigo1999ant,Dorigo1999}. Basic algorithm of ACO is based on laying of pheromone trail inspired by biological ants’ behavior, which use pheromones as medium of communication. In a similar way to the biological example, simple agents – artificial ants – communicating with each other indirectly using artificial pheromones within a colony is base of ACO. ACO is based on indirect communication within a colony of simple agents, called ants (artificial), induced by (artificial) pheromone bands. Traces of pheromones work as distributed and digital information in ACO which ants use to build solutions to the problem to be solved and to adapt to it during the execution of the algorithm as a reflection of their search experience which is a major difference between ACO algorithm and other construction heuristics.
Ant System (AS) ~\cite{dorigo1992optimization,dorigo1991ants,dorigo1996ant69} was one of the starting example of such algorithm that was put forward using an eminent application example – Travelling Salesman Problem (TSP) ~\cite{applegate2006traveling6,reinelt1994traveling155}. Though Ant System presented promising results but it was still not enough to compete with the most advanced TSP algorithms. Still, it played an important role to encourage further exploration of the two algorithmic variants that provided significantly better computing power and applications. In fact, quite astonishing results are achieved in significantly large number of applications using this algorithm. Problems like DNA sequencing ~\cite{blum2008ant27}, scheduling ~\cite{blum2005beam20}, protein-ligand docking ~\cite{korb2007ant107}, assembly line balancing ~\cite{blum2008beam21}, sequential ordering ~\cite{gambardella2000ant84}, packet-switched routing ~\cite{di1998antnet52} and 2D-HP protein folding ~\cite{shmygelska2005ant160} are some of example applications of ACO.  A common framework is provided by ACO metaheuristic for algorithmic variations and prevailing applications ~\cite{dorigo1999ant}. ACO metaheuristic following algorithms are called ACO algorithms. 
Hundreds of researchers have been attracted towards with ACO field since early version of ACO algorithm was proposed in 1991. ACO is now considered as a heavily developed metaheuristic and there exists a great number of both theoretical and experimental research results by now. ACO importance can be demonstrated by ~\cite{doerner2009special57}:
\begin{enumerate}
\item	IEEE Swarm Intelligence Symposium series
\item	Biannual International Conference on Ant Colony Optimization and Swarm Intelligence (ANTS)
\item	Several conferences on evolutionary algorithms and metaheuristics
\item	Special journal issues
\item	Dedicated website and mailing list for ACO (www.aco-metaheuristic.org)
\end{enumerate}

The rest of this article is divided into four sections. The first section, presents a brief background of ACO algorithm followed by initial developments performed in ACO algorithm in the second section. The third section, gives a detailed overview of recent developments (after 2010) in terms of application development, algorithmic development and parallel implementation. The fourth and last section concludes this article. 

\section{Historical Background}
Ant System (AS) was preliminary proposed ACO algorithm that was applied to relatively lesser TSP instances not more than 80 cities. Performance matching some other common heuristics like evolutionary computation ~\cite{dorigo1992optimization,dorigo1996ant69} was achieved by initial AS yet it was not enough to match or compete performance given by other advanced algorithms designed for TSP. As a result, focus of research within the ACO has been shifted towards ACO algorithms that work better than AS, for example when applied to the TSP. The rest of this section, provides introduction of biological metaphor that was inspiration for both AS and ACO. 
\subsection{Biological Analogy}
In most ant species, a chemical known as pheromone is deposited on ground by ants while walking which is smelled by other ants ~\cite{deneubourg1990self48}. From nest to food source, a path is marked through a trail created due to pheromones deposited by ants. Other ants in search of food, use the pheromone trail (path) to locate the food. Many ant species are proficient in determining shortest trail from nest to food location by exploiting pheromone trails. Deneubourg et al ~\cite{deneubourg1990self48}, created a controlled experimental environment to study and research ants behavior in pheromone laying and determination of shortest path to food. For this purpose, they used a variable length double bridge to connect food source and nest of ants. They ran multiple experiments with one longer bridge and other shorter bridge. Results showed that, in the beginning of each experiment, ants started a random motion towards the food in the arena but shorter path was used by ants in the end. Results presented by Deneubourg et al ~\cite{deneubourg1990self48}, can be explained as follows. In the start, no ant travelled the environment therefore there was pheromone present on the ground which gives same probability of any route selection without any preference. It can be said that, on average nearly half number of ants chose each path in the beginning. However, those ants which selected shorter path reached the food source and travelled back to nest earlier than other set of ants which selected longer path. Now, to again travel back towards food source, ants decide to choose one path and due to presence of higher pheromone level on shorter path prejudice their decision. In the end, pheromone level on shorter path start to add up faster causing nearly entire colony of ants to use that path. 
In Ant Systems and different variants of ACO algorithms, same phenomenon is applied. Ants are replaced by artificial ants (agents). Artificial pheromone trails are used to represent pheromone trails and graph is used to represent double bridge. Extra capabilities to implement restrictions and retracing of solution without errors are given to artificial ants to solve real life problems. Quantity of pheromone deposited by artificial ants is set at proportional to quality of produced solution – a behavior similar to biological ants ~\cite{beckers1993modulation10}.

\section{Initial Developments}
Marco Dorigo was first person to propose and publish first ACO algorithm – Ant System (AS) – in a set of three algorithms namely ant-quantity, ant-density, and ant-cycle as part of his doctoral thesis ~\cite{dorigo1992optimization}. Few years later, these three algorithm appeared first as technical report ~\cite{dorigo1991ants} in the IEE Transactions on Man, Cybernetics and Systems ~\cite{dorigo1996ant69}. Difference between three algorithms was that in ant-quantity and ant-density, pheromone was updated right after ants move from one city to another, while in ant-cycle, pheromone deposition was updated once all ants had built the path and tour quality was used as a function for level of pheromone update. Better performance presented by ant-cycle caused research in other two algorithms to stop and ant-cycle was used to present Ant System. Initial version of AS, presented encouraging results but were not enough to compete with other well-established algorithms. However, these results were encouraging enough to stimulate exploration of research in this field.
\subsection{ACO Algorithm}
Basic ACO algorithm is given in Figure \ref{ACOAlgo} and consists of four main stages. Each stage is explained as follow:
\begin {itemize}
\item \textbf{Initialization}: This is first step, in which all parameters and pheromones variables are set
\item \textbf{Construct Ant Solutions}: After initialization, a set of ants build solution to the problem being solved using pheromones values and other information
\item \textbf{Local Search}: This stage is optional and it includes improvement of constructed solution by ants
\item \textbf{Global Pheromone Update}: This is last stage, and it includes update in pheromone variables based on search experience reflected by ants.
\end {itemize}

\begin{algorithm}[H]
	\textbf{procedure} \textit{Start of ACO Algorithm}\\
	initialization\;
	\While{(Iterate until end criteria is reached)}{
		BuildAntSolutions\;
		LocalSearchMethod  \% optional \;
		PheromonesGlobalUpdate\;
	}
\textbf{end} \textit{End of ACO algorithm}
	\caption{ACO Algorithm}
	\label{ACOAlgo}
\end{algorithm}

\subsection{Earlier Variants of ACO}
There have been numerous updates and improvements made in ACO algorithms which are presented and published by many researchers in the literature. A brief overview on each variant can take a lot time to report therefore this article only focuses on details of development done in ACO algorithm after year 2010. Earlier variants of ACO algorithms with proposal year and main references are presented in Table \ref{ealierDev}.

\begin{table}
	\centering
	\caption{Earlier Developments in ACO Algorithm}
	\label{ealierDev}
\begin{tabular}{|c|c|c|}
	\hline 
\textbf{	ACO Algorithm} & \textbf{Year} & \textbf{Main Reference} \\ 
	\hline 
	Ant System & 1991 & ~\cite{dorigo1992optimization,dorigo1991ants,dorigo1996ant69} \\ 
	\hline 
	Elitist AS & 1992 & ~\cite{dorigo1992optimization,dorigo1991ants,dorigo1996ant69} \\ 
	\hline 
	Ant-Q & 1995 &  ~\cite{gambardella1995ant82} \\ 
	\hline 
	Ant Colony System & 1996 &  ~\cite{dorrigo1997ant64} \\ 
	\hline 
	Max-Min Ant System & 1996 & ~\cite{stutzle1996improving174,stutzle1997max175} \\ 
	\hline 
	Rank-based AS & 1997 & ~\cite{bullnheimer1997new31} \\ 
	\hline 
	ANTS & 1998 & ~\cite{maniezzo1999exact124} \\ 
	\hline 
	Best-worst AS & 2000 & ~\cite{cordon2002analysis38,cordon2000new39} \\ 
	\hline 
	Population based ACO & 2002 & ~\cite{guntsch2002population92} \\ 
	\hline 
	Beam-ACO & 2004 & ~\cite{blum2004theoretical19,blum2005beam20} \\ 
	\hline 
	Hyper Cube – ACO & 2004 & ~\cite{blum2001hcHCACO} \\ 
	\hline 
\end{tabular}
\end{table} 

\section{Developments}
As mentioned earlier, covering updates and improvements in ACO algorithm from initial proposed system will take a lot of time, therefore, this review article covers recent evolutions in ACO algorithms after year 2010. In this section, latest research updates in ACO are presented. This section is divided into three sub sections. The first subsection, outlines application of Ant Colony Optimization algorithms to non-standard problems followed by ACO developments from algorithmic point of view in the second subsection. The third and the final subsection presents improvements in ACO algorithm with parallel implementations.
\subsection{ACO Non-standard Applications}
This subsection presents a review of ACO application to problems involving factors such as time-varying data, multi-objective function and continuous optimization.
\subsubsection{NP-hard Problems}
Dynamic problems where characteristics change during runtime such as network routing are main applications in which ACO algorithm have been significantly successful. Classical NP-hard problems with dynamic variants is one of main area where ACO algorithms have also been applied. One such example is dynamic version of Travelling Salesman Problem (TSP) where cities may disappear or appear or distance between cities may change. Mavrovouniotis et al ~\cite{mavrovouniotis2017ant133} presented a more recent work in this field that targeted towards improvement in performance of ACO on dynamic problems by explicitly using local search algorithms. Mavrovouniotis and Yang in 2015 ~\cite{mavrovouniotis2015ant131} reported ACO algorithm application to dynamic vehicle routing problems with better results on both real-world instances and academic instances. Mavrovouniotis et al ~\cite{mavrovouniotis2017survey132} published another review article on dynamic optimization problems with swarm intelligence algorithms such as ACO. 
\subsubsection{Multi-Objective Optimization}
In real-world applications, multiple solutions are assessed as a function of several, mostly differing objectives. Different approaches are used to handle these problems. One approach is to order objectives according to their importance as reported by Veen et al ~\cite{van2013ant85} as part of their research. They presented a two-colony Ant Colony System (ACS) algorithm – MACS-VRPTW – to solve routing problem of vehicles with time windows. López-Ibáñez and Stützle proposed a different approach to develop multi-objective ACO algorithms ~\cite{lopez2012experimental119}. Different existing methods to manage multi-objective problems were analyzed by them and a generic multi-objective ACO version (MO-ACO) is proposed. This version can be used to instantiate other approaches as well as new variations can also be developed. By exploring the design space created by MOACO algorithms with the new methodology of creating multi-objective optimizers, new MOACO algorithms were automatically generated that outperformed all the earlier proposed ACO algorithms in this field ~\cite{lopez2012automatic118}. In more recent research, Falcón-Cardona and Coello Coello further extended the framework presented by Stützle et al to present a novel approach for multi-objective problems with ACO algorithm variant called iMOACO-RR ~\cite{falcon2017new77}.
\subsubsection{Continuous Optimization}
Initially ACO was applied to combinatorial problems but later on it was adapted for continuous optimization problems. Variables’ discretization of real-valued domain is simplest methodology to apply ACO to continuous problems. Implementation of ACO using this approach has been applied to Protein-ligand docking problem. Socha and Dorigo first presented an extended version of ACO named as ACOR algorithm where Gaussian kernel functions are explicitly used in place of probability density functions. Liao et al presented a refined form of ACOR algorithm using growing population size and incorporating dominant local search algorithms ~\cite{liao2011incremental113}. Kumar et al ~\cite{kumar2015enhancing109} and Guo et al ~\cite{guo2012antNew}, later on presented refined forms of same algorithm. Liao et al ~\cite{liao2014unified114} presented an integrated structure for ACO applications to continuous optimization. This version of ACO allows selection of particular algorithm components to instantiate different variants of ACOR. Unified structure was able to generate ACO algorithms using Irace – an automatic algorithm setting tool, that outperformed all the earlier versions presented in the literature ~\cite{lopez2016irace122}. Yang et al ~\cite{yang2017adaptive187}, also presented an extended version of ACOR for multi-modal optimization.
\subsection{Developments in Algorithms}
This sub section presents review of research articles that mainly focused on development of ACO variants to improve performance. This includes ACO hybridization with other algorithms and inter-programming techniques. 
\subsubsection{ACO Combination with Other Metaheuristics}
To fine-tune solutions built by ants, local improvement heuristics are combined with ACO as most common technique. For this purpose, repetitive improvement algorithms are mainly used but many other articles have reported use of other metaheuristic algorithms for improvement. Oliveira et al ~\cite{oliveira2017analysis176} presented use of tabu search for quadratic assignment problem in order to improve solution constructed by ants. Many other hybridization techniques have been proposed such as letting ants to build a solution from partially present solution as reported by Stutzle et al ~\cite{stutzle2018iterated185} in their article in the form of iterated ants. Another approach presented by Hara et al ~\cite{hara2010ant182} is to use other complete solutions to take out partial solutions. Using former approach, fasten the process of solution construction as well allows exploitation of good parts of solution directly. Results presented by researchers using this approach showed that this approach is quite effective in the absence of other useful local search methods. 
\subsubsection{ACO Combination with Inter-Programming Techniques}
Integration of ACO with programming techniques is very useful for highly constrained problems for which finding a feasible solution is not so easy. Blum et al ~\cite{blum2011hybrid139} presented an approach that require integration of ants’ solutions construction with constraint propagation mechanism to determine if certain solution will be feasible or not. This approach have shown good results for scheduling problems that are highly constrained.Solnon et al ~\cite{solnon2013ant105}, recently published his research that focuses on integration of constraint solver into ACO algorithm. Column generation technique is integrated with ACO to solve vehicle routing problems that include black-box achievability constraints is presented by Massen et al ~\cite{massen2012pheromone128}. This research was further explored and tested on various benchmarks with improved results in extended version of research article by Massen et al ~\cite{massen2013experimental129}.
\subsection{Parallel Implementations}
The nature of the ACO algorithms allows parallelization to be applied to both population and data domain. Other algorithms based on population using parallelization schemes can be upgraded to ACO. Course grained and fine grained techniques are used to as classification of older parallelization strategies. Fine parallelization is characterized by the fact that very few people are assigned to a single processor and that information is exchanged frequently between processors. On the other hand, coarse-grained approaches use bigger population sometime entire population to be attributed to individual processor and the exchange of information is rather rare. Merkle et al ~\cite{merkle2002modeling134} provides a brief overview of both approaches.
Fine-grained parallelism schemes were explored very early, when shared memory models and multicore CPUs were not common or unavailable. Initial fine-grained strategy were investigated with parallel AS versions for the TSP on the CM-2 Connection Machine by assigning each Ant to a single processing unit. However, results presented by experiments using this approach reported negative values due to communication overhead between ants for modification of pheromone trails. On other hand, research have showed that coarse-grained strategies are much more effective for ACO. For implementing course-grained parallel technique to ACO, p sub-colonies are executed in parallel depending on p available processors. Uchida et al ~\cite{uchida2012efficient123} used this technique to present results that proved running p independent sub-colonies an effective technique with further improvements achieved by improving exchange of information among sub-colonies. These improvements were studied and presented in detail by Twomey et al ~\cite{twomey2010analysis184} in their research. Presence of shared memory models and multicore CPU architecture allows introduction of thread level parallelism into ACO algorithm that can further improve speed of ACO algorithm. Guerrero et al ~\cite{guerrero2014comparative90} evaluated parallel versions of ACO algorithms on various platforms while most recent trends have seen implementation of ACO algorithms on GPUs to speed-up, as reported by ~\cite{cecilia2013enhancing35,dawson2013improving43,delevacq2013parallel46}.  
\section{Conclusion}
Since its introduction in 1991 by Dorigo, ACO algorithm has become one of most widely used metaheuristic for solving combinatorial problems. Started as Ant System (AS), initial versions of ACO were not promising enough to compete with other established algorithms but results were encouraging enough to open up the gates of research in this field. Since then, many researchers have exploited basic version of ACO to work on bringing updates in it for achieving better results. This research focused on outlining more recent developments carried out in ACO algorithms in terms of both applications and algorithmic developments. For non-standard applications of ACO algorithms, multi-objective optimization, continuous optimization and time-varying NP-hard problems were main focus of recent developments. While to improve, performance of ACO algorithm, it has been integrated with other metaheuristics and inter-programming techniques. Hybridization of ACO algorithms have clearly showed a lot of improvement in results for various problems. Recent trends have also seen implementation of ACO algorithms with parallel versions. Availability of multicore CPU architectures and GPUs have made it possible to develop improved parallel versions of ACO algorithms.  

\bibliographystyle{unsrt}


\bibliography{acobib}

\end{document}